\documentclass[a4paper,conference]{IEEEtran}
\ifCLASSINFOpdf
\else
\fi

\usepackage{graphicx}
\usepackage{amsmath}
\usepackage{amsthm}
\usepackage{booktabs}
\usepackage{algorithm}
\usepackage{algorithmic}
\usepackage{epsfig}
\usepackage{amssymb}
\usepackage{multirow}
\usepackage{float}
\usepackage{wrapfig}
\usepackage{dsfont}
\usepackage{multicol}
\usepackage{color}
\usepackage{url}
\renewcommand{\baselinestretch}{0.982} 

\begin{document}
%
\title{Unsupervised Domain Adaptation for Video Transformers in Action Recognition}

\author{Victor G. Turrisi da Costa$^1$, Giacomo Zara$^1$, Paolo Rota$^1$, Thiago Oliveira-Santos$^2$, \\ Nicu Sebe$^1$, Vittorio Murino$^{3,4}$ and Elisa Ricci$^{1,5}$ \\ 
$^1$University of Trento, $^2$Universidade Federal do Espírito Santo, $^3$University of Verona, \\ $^4$Pattern Analysis \& Computer Vision (PAVIS) - Istituto Italiano di Tecnologia, $^5$Fondazione Bruno Kessler \\
{\tt\small \{vg.turrisidacosta,giacomo.zara,paolo.rota,niculae.sebe,e.ricci\}@unitn.it} \\ {\tt\small todsantos@inf.ufes.br}, {\tt\small vittorio.murino@iit.it}
}

\maketitle

\begin{abstract}
Over the last few years, Unsupervised Domain Adaptation (UDA) techniques have acquired remarkable importance and popularity in computer vision.
However, when compared to the extensive literature available for images, the field of videos is still relatively unexplored.
On the other hand, the performance of a model in action recognition is heavily affected by domain shift.
In this paper, we propose a simple and novel UDA approach for video action recognition. Our approach leverages recent advances on spatio-temporal transformers to build a robust source model that better generalises to the target domain.
Furthermore, our architecture learns domain invariant features thanks to the introduction of a novel alignment loss term derived from the Information Bottleneck principle.
We report results on two video action recognition benchmarks for UDA, showing state-of-the-art performance on \textit{HMDB$\leftrightarrow$UCF}, as well as on \textit{Kinetics$\rightarrow$NEC-Drone}, which is more challenging.
This demonstrates the effectiveness of our method in handling different levels of domain shift.
The source code is available at \url{https://github.com/vturrisi/UDAVT}.
\end{abstract}


%
\IEEEpeerreviewmaketitle

\section{Introduction}
\label{sec:intro}

The standard setting for visual learning tasks relies on the assumption that training and testing data belong to the same domain, i.e., they are drawn from the same distribution.
However, in practice, this often does not hold, as many real-world scenarios require models to be tested on very different, yet related, domains.
This often leads to poor performance in the domain of interest, especially when the domain shift is significant.
To address this issue, Unsupervised Domain Adaptation (UDA) techniques have been developed with several approaches achieving remarkable results on image-related tasks.
However, much less attention has been devoted to videos, which poses a greater challenge given the significant increase in complexity that arises when the temporal aspects come into play.

Action recognition \cite{feichtenhofer2016convolutional,zhou2018temporal,tran2015learning,carreira2017quo} is one of the most popular tasks when it comes to video analysis.
This task is particularly challenging due to some inherent characteristics of sequential visual data, mostly related to the significant variability that arises due to the many different ways in which a certain human activity can be carried out, with variations in terms of speed, duration, relative movement between camera and actor(s), occlusion, etc.
These issues, along with many others, have been tackled in several previous works \cite{feichtenhofer2016convolutional,zhou2018temporal,tran2015learning,carreira2017quo} that demonstrated remarkable performance exploiting different deep architectures. Nevertheless, many open problems still exist when domain shift arises in video analysis. 

\begin{figure*}[!t]%
    \centering
    \centering \footnotesize{\textbf{(a)} Source only} \\
    {{\includegraphics[width=0.6\textwidth]{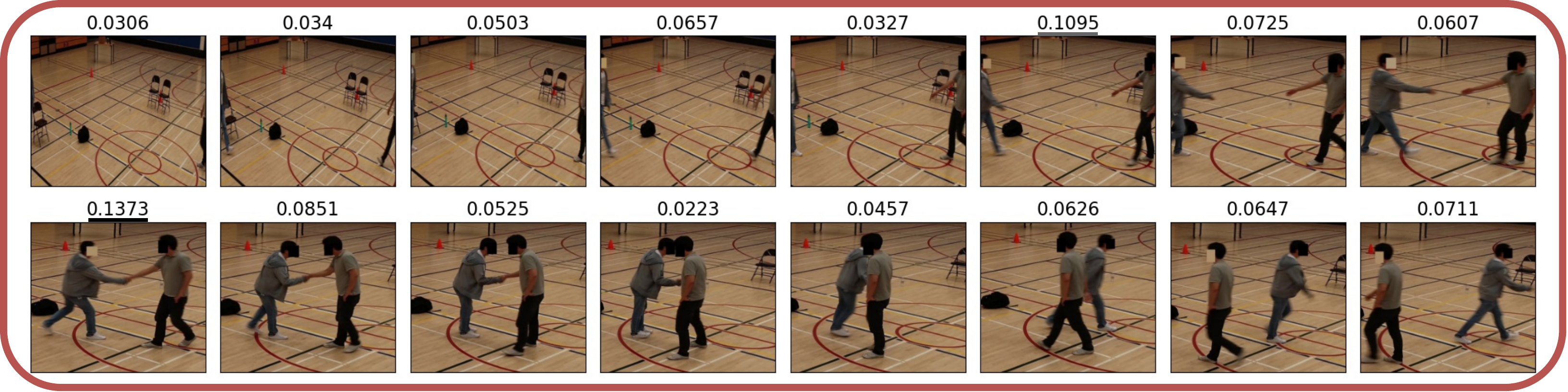} }}%
    \\
    \centering \footnotesize{\textbf{(b)} IB Domain adaptation} \\
    {{\includegraphics[width=0.6\textwidth]{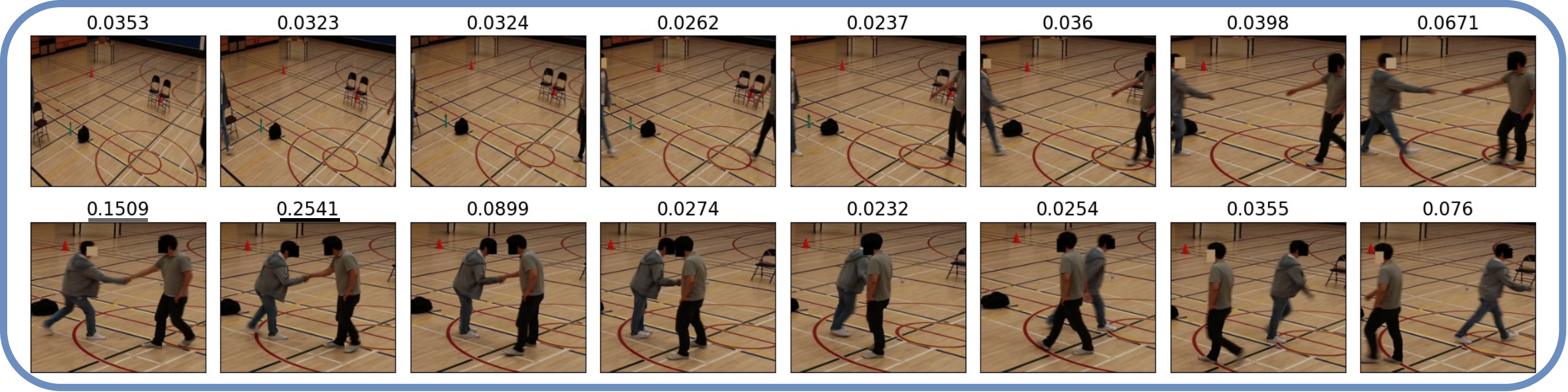} }}%
    \\
    \caption{Temporal attention visualisation: comparison between a source-only model ({a}) and UDAVT ({b}). Most and second most attended frames are highlighted.} 
    \label{fig:attention}
    \vspace{-12pt}
\end{figure*}

In this work, we tackle the problem of UDA in video action recognition by proposing a novel approach that exploits a new information theoretic-based domain alignment loss on top of a transformer-based feature extractor. 
In particular, our architecture is built upon a shared encoder derived from the STAM \cite{sharir2021image} visual transformer coupled with a domain alignment loss inspired by the Information Bottleneck (IB) principle \cite{tishby2000information,tishby2015deep}. The encoder is composed of a spatial transformer, e.g. ViT \cite{dosovitskiy2021image}, that processes individual frames and a temporal transformer, e.g. a simple multi-layer transformer as in \cite{vaswani2017attention}, that aggregates spatial features to process a whole video. In this paper, we propose a two-phase process for training, where we first fine-tune the whole transformer using only source data as in \cite{lu2021pretrained}, and then we freeze the spatial transformer and fine-tune only the temporal one with the novel IB loss, which promotes domain distribution alignment. The seamless integration of the transformer with our adaptation strategy leads to improved recognition accuracy (see Sec. \ref{experiments}) that stems from a better spatio-temporal action localisation.
As an example, Fig. \ref{fig:attention} depicts a 16-frame sequence and the associated frame-level attention values for a validation video, comparing the model trained only on source data ({a}) and the same model trained with our UDA approach ({b}).
From the picture, it is possible to see that our method enables the network to better focus on the frames where the handshake action takes place. We called our method UDAVT (\textit{Unsupervised Domain Adaptation for Video Transformers in Action Recognition}). We evaluate our approach on two popular UDA benchmarks for action recognition, namely \textit{HMDB$\leftrightarrow$UCF} \cite{chen2019temporal} and \textit{Kinetics$\rightarrow$NEC-Drone} \cite{choi2020unsupervised}, and show that UDAVT outperforms previous UDA methods.

\noindent \textbf{Contributions.} Our contributions are as follows: (i) we propose a novel UDA approach for video action recognition that exploits class label information within a novel IB domain alignment loss; (ii) we show, for the first time, that a transformer-based feature extractor can be successfully employed in UDA for building a robust source model that is more resilient to domain shift than traditional architectures for video analysis; (iii) we provide an extensive evaluation of our method and show that UDAVT outperforms state-of-the-art UDA approaches on all considered datasets.

\section{Related Work}
\label{sec:related}

\noindent \textbf{Action recognition.} In the last few years, several approaches and architectures have been proposed for video action recognition.
For instance, two-stream networks were proposed in \cite{feichtenhofer2016convolutional}, jointly exploiting RGB and optical flow sequences.
The Temporal Relation Network was proposed by Zhou \textit{et al.} \cite{zhou2018temporal} to model temporal relations across the sequence by employing a specific pooling layer.
Other works opted for 3D CNNs to learn spatio-temporal features \cite{tran2015learning, carreira2017quo}.
Recently, some works proposed contrastive-learning-based methods in order to extract efficient motion representations for action recognition \cite{yang2020hierarchical,qian2020spatiotemporal,wang2020self,patrick2020multi}.
In order to model the temporal aspects, fusion methods have been proposed \cite{meng2021adafuse,afza2021}, as well as, combinations of recurrent and convolutional networks \cite{Stergiou_2021}.
Bai \textit{et al.} \cite{bai_interchange} introduced an approach based on multi-range feature interchange to capture short-range motion features and long-range dependencies.
Finally, advanced transformer-based approaches have been proposed for action recognition \cite{girdhar2019video,sharir2021image, bertasius2021spacetime, huang2021training, 2022acttrans, neimark2021video}, often exploiting skeleton points \cite{10.1007/978-3-030-68796-0_50, 9312201, s21165339, 9428459, 10.1145/3444685.3446289}. 
Different from our work, these approaches consider a standard supervised setting, where all data is labelled and there is no domain shift between training and evaluation datasets. 

\noindent \textbf{UDA for images.} Existing UDA methods differ from the strategy adopted to address the domain shift.
One option consists of matching statistical moments of the source and target data distributions \cite{long2015learning,carlucci2017autodial,roy2019unsupervised,mancini2018boosting,tzeng2014deep}, eventually integrating label information in the alignment loss \cite{kang2019contrastive}.   
Another common approach is based on adversarial learning \cite{long2018conditional,hong2018conditional,ganin2016domain,Hoffman:Adda:CVPR17} and aims at learning discriminative and domain-agnostic feature representations by combining implementing a domain discriminator.
Another possibility is to use Generative Adversarial Networks \cite{hoffman2017cycada,volpi2018adversarial,sankaranarayanan2018generate} to generate target-like instances from source data that are then employed to train the model.
Furthermore, self-supervised learning has been recently exploited for DA, in particular when devising auxiliary tasks, such as predicting rotations \cite{sun2019unsupervised} or image patches permutations \cite{bucci2019tackling}, to learn domain-invariant feature representations.
Finally, recent works proposed to employ transformer-based architectures in the scope of domain adaptation \cite{yang2021transformerbased, xu2021cdtrans}.
However, all these methods were proposed for images.

\noindent \textbf{UDA for action recognition} The challenging problem of domain shift for video action recognition has been addressed by a limited number of works \cite{munro2020multi,pan2020adversarial,chen2019temporal,choi2020unsupervised,Choi2020ShuffleAA}, despite the several possible applications it finds in real-world scenarios. Chen \textit{et al.} \cite{chen2019temporal} proposed TA$^3$N (\textit{Temporal Attentive Adversarial Adaptation Network}), which employs a temporal relation module to jointly perform alignment between the source and target domains and learn the temporal relation across the video sequences.
Pan \textit{et al.} \cite{pan2020adversarial} proposed TCoN (\textit{Temporal Co-attention Network}), a deep architecture integrating a cross-domain attention module in order to match the distributions of source and target domain between temporally aligned feature representations.
Choi \textit{et al.} proposed two methods based on adversarial learning \cite{choi2020unsupervised} and on an attention mechanism \cite{Choi2020ShuffleAA}, respectively.
Kim \textit{et al.} \cite{Kim_2021_ICCV} proposed an UDA approach based on learning cross-modal contrastive features, while in \cite{munro2020multi} an approach was introduced based on self-supervised and multimodal learning (RGB + optical flow).
Song \textit{et al.} \cite{Song_2021_CVPR} proposed a method for spatio-temporal contrastive domain adaptation.
Lastly, Turrisi da Costa \textit{et al.} \cite{da_Costa_2022_WACV} proposed CO$^2$A, an approach for UDA that relies on a contrastive loss to perform domain alignment, in addition to stabilising training by making a classification and a contrastive head agree.
Our proposed approach presents similarities with previous self-supervised based frameworks \cite{park2020joint,kim2020cross}, but introduces novelty in employing an Information Bottleneck loss to align domains rather than to learn better feature representations.
Additionally, our work is also the first to exploit transformers for UDA in the context of action recognition.
\section{Proposed method}
\label{sec:method}

\noindent \textbf{Problem setting and notation.}
The paper tackles the problem of UDA for action recognition. Given a source dataset $\mathcal{S} = \{X^S_i, y^S_i\}_{i=1}^{N_S}$ of videos and associated annotations, and an unlabelled target dataset $\mathcal{T} = \{X_i^T\}_{i=1}^{N_T}$, where $X_i \in \mathcal{X}$ and $y \in \mathcal{Y}$, $\mathcal{Y}=\{1,2, \dots, K\}$ ($K$ denotes the number of action categories), we aim to learn a function $F_\theta: \mathcal{X} \rightarrow \mathcal{Y}$ with parameters $\theta$ that maps an input video $X$ to a class label $y$ and perform well on target data. 
Note that this is not a trivial task, since source and target data are sampled from two different distributions, $P^S(X) \neq P^T(X)$.
To tackle this problem, we propose a novel approach called UDAVT that combines two main components: a spatio-temporal transformer architecture and a novel distribution alignment scheme derived from the IB principle \cite{tishby2000information}.   

\begin{figure*}[t]
    \centering
    \includegraphics[width=\textwidth]{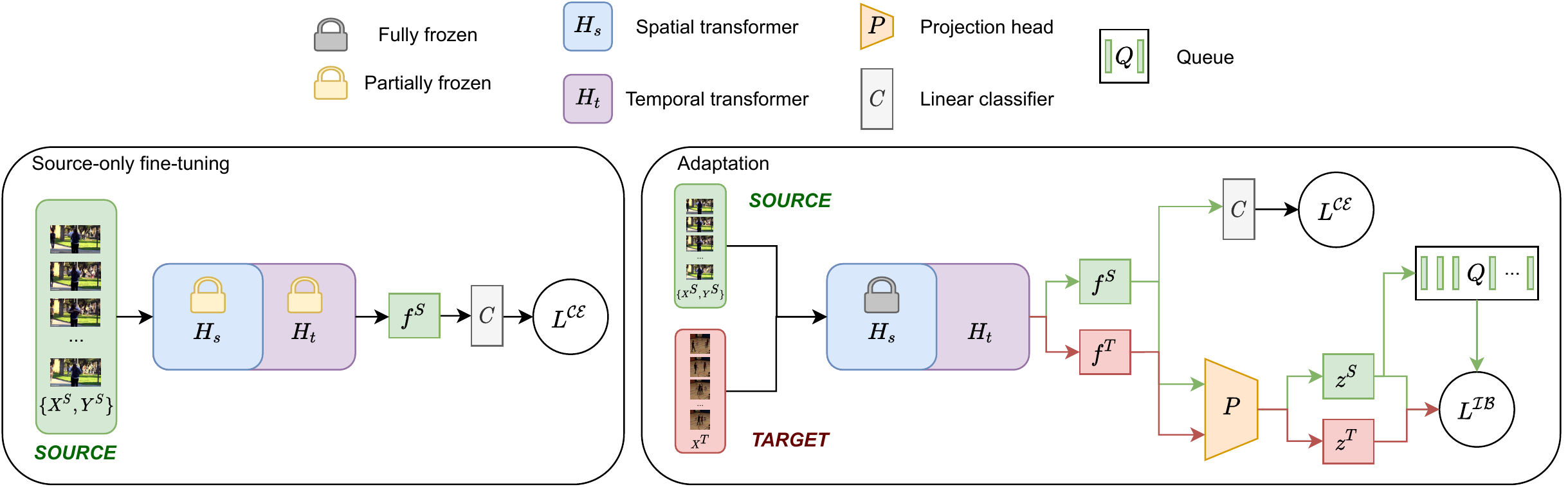}
    \vspace{-10pt}
    \caption{\textbf{Overview of UDAVT.} Our approach is articulated in two steps. In phase 1 (left), source data are fed to a video transformer $H$ (composed of a spatial transformer $H_s$ and a temporal transformer $H_t$) followed by a classifier $C$. The overall model is finetuned with a supervised cross-entropy loss $L^\mathcal{CE}$ similarly to \cite{lu2021pretrained}. In phase 2 (right), the weights of $H_s$ are frozen, while $H_t$ is fine-tuned. Source and target data are fed to the backbone and the proposed IB-based loss $L^\mathcal{IB}$ performs domain alignment, while $L^\mathcal{CE}$ further trains the action classifier. A queue $Q$ is added in order to increase the number of source instances considered while computing $L^\mathcal{IB}$.}
    \vspace{-10pt}
    \label{fig:overview}
\end{figure*}

\noindent \textbf{Overview.} An overview of UDAVT is shown in Fig.\ref{fig:overview}. We propose a two-phase training pipeline where the model is first trained with source data and subsequently adapted using source and target data. Our model is defined as $F_\theta=C \circ H$, where $H$ represents a video transformer encoder \cite{sharir2021image} and $C$ represents a linear classifier.
$H$ is composed by two main parts, a spatial transformer $H_s$ that extracts frame-level feature representations and a temporal transformer $H_t$ that aggregates the frame-level features to produce video-level representations. In particular, $H_s$ is the vision transformer ViT \cite{dosovitskiy2021image}, whereas $H_t$ is a simple multi-layer transformer as in \cite{vaswani2017attention}.
An auxiliary MLP projection head $P$ is also used in the second phase.
Finally, the complete model also has a queue $Q$ that is responsible for keeping the most recent feature representations of source data.
The two main phases of our approach are described as follows.

\noindent \textbf{Phase I: Source-only fine-tuning.} The training process of this phase starts from a model ${H}$ pretrained on the Kinetics dataset~\cite{kay2017kinetics} and consists of fine-tuning the entire model $F_\theta$ using only the source data $\mathcal{S}$. As in \cite{sharir2021image}, we consider 16 frames uniformly sampled to represent a source video $X^S$ as input to $F_\theta$. The first part of the model $H_s$ divides each frame into 16x16 patches that are then projected into feature vectors.
$H_s$ consists of ViT, which receives as input the projected patches together with a classification token $[CLS]^S$ as in \cite{devlin2019bert}.
Each frame is processed individually, extracting feature representations $f_{{H}_s}^S = [CLS]^S$ that consists of the classification token linked to that specific frame.
During the forward pass, $[CLS]^S$ will collect all important information from the image patches.
The frame-level features $f_{{H}_s}^S$ are then forwarded through ${H}_t$ together with a new classification token $[CLS]^T$ which, after processing, produces the video-level feature representations $f^S = [CLS]^T$.
In this step, ${H}_s$ and ${H}_t$ are fine-tuned following the strategy proposed in \cite{lu2021pretrained}, which consists of freezing all parameters except the positional encoding, the input embeddings, the classification tokens and the affine transformations inside the layer normalisations \cite{ba2016layer}.
While \cite{lu2021pretrained} studied the problem of partially fine-tuning a transformer for handling different modalities, in this work, we show that this strategy can be successfully applied to the problem of domain adaptation.
Finally, the video level features are then fed to a linear classifier $C$. The entire model is trained with a supervised cross-entropy loss $L^{\mathcal{CE}}$, defined as:
\begin{equation}
    L^{\mathcal{CE}} = - \mathbb{E}_{(X, y) \, \in \, \mathcal{S}} \sum y_k \, \log \, \sigma(F_\theta(X)),
\end{equation}
where $\sigma$ is the softmax operation. Due to lack of space, the reader is referred to \cite{sharir2021image} and \cite{dosovitskiy2021image} for more details about the transformer architecture.

\begin{figure}[t]
    \centering
    \includegraphics[width=0.25\textwidth]{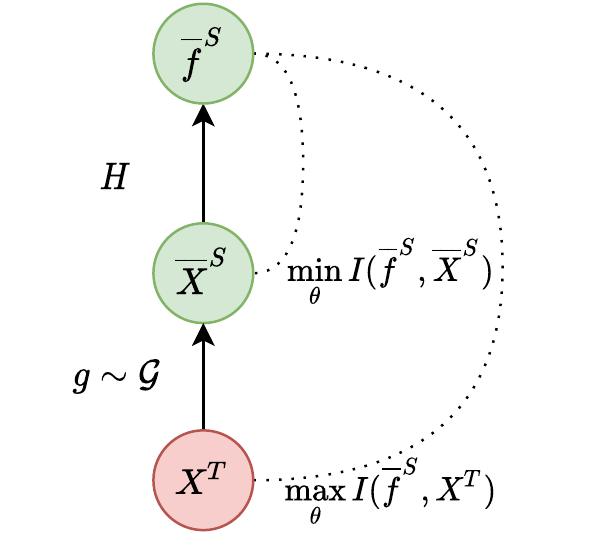}
    \caption{Information Bottleneck diagram showing the proposed flow of information to perform adaptation.}
    \label{fig:mutual-info}
    \vspace{-10pt}
\end{figure}

\noindent \textbf{Phase II: Target Adaptation.} In this phase, the spatial transformer ${H}_s$ is frozen, while the parameters of ${H}_t$ are trained to exploit both labelled source and unlabelled target data.
Note that both ${H}_s$ and ${H}_t$ are initialised with the weights after Phase I.
This choice is motivated by the need of reducing computational resources, while still performing adaptation at the temporal level. Freezing part of the model enables us to increase the batch size, which is fundamental for the proposed domain alignment strategy (Eqn. \ref{eq:IB}). 
To train our model, 16 frames are sampled from both source $X^S$ and target $X^T$ videos, as in the previous phase. Video-level feature representations $f^S$ and $f^T$ are then produced for videos of both domains.
Subsequently, the temporal features of source videos $f^S$ are provided as input to the linear classifier $C$.
To perform adaptation, we rely on the Information Bottleneck (IB) principle \cite{tishby2000information,tishby2015deep}.
Fig. \ref{fig:mutual-info} shows how the IB principle is applied to our problem. 
First, we assume that there exists a domain transformation $g \sim \mathcal{G}$ that maps a target instance $X^T$ to a source instance $\overline{X}^S$ that has the same label. 
Unlike \cite{zbontar2021barlow}, which considers that one instance is mapped to a perturbed version of the same instance via some type of data augmentation, we map a single target instance $X^T$ to multiple different $\overline{X}^S$ in the same iteration.
We experimentally show that this is indeed beneficial since by increasing the number of source instances via the usage of a queue, and consequently the number of pairs, we observed a large boost in performance.
As annotations are not provided for the target domain, we resort to pseudo-labels for matching source and target instances.
The model $H$ maps $\overline{X}^S$ to the feature representation $\overline{f}^S$. According to the IB principle, we want the model $H$ to learn a representation $\overline{f}^S$ which encodes as much information as possible about the original instance $X^T$. This objective is carried out by maximising the Mutual Information $I(\overline{f}^S, X^T)$. 
Then, the second objective consists of minimising $I(\overline{f}^S, \overline{X}^S)$ to make the model $H$ invariant to the transformation of the sample $X^T$ into a different domain.
The overall loss function can be written as:
\begin{equation}
    {L}^{\mathcal{IB}} = I(\overline{f}^S, \overline{X}^S)- \beta \, I(\overline{f}^S, X^T)
    \label{eq:info-bottleneck}
\end{equation}

Since optimising for mutual information in a high dimensional space is difficult, previous works have proposed different ways to approximate Eqn. \ref{eq:info-bottleneck}. In this work, we derive a loss function similar to that used in the Barlow Twins method \cite{zbontar2021barlow}, where it was proved that, under certain conditions, Eqn. \ref{eq:info-bottleneck} can be approximated as:
\begin{equation}
\begin{aligned}
{L}^{\mathcal{IB}} = \sum_i^d (1 - C_{ii})^2 + \lambda \, \sum_i^d \sum_{j \neq i}^d (C_{ij})^2,
\end{aligned}
\label{eq:IB}
\end{equation}
where $\boldsymbol{C}$ is a cross-correlation matrix computed over a batch of $B$ data obtained through a feature extractor $\{z_1, \dots, z_B\}$ and their
corresponding transformed version $\{z_1', \dots, z_B'\}$, where $i$ is the feature index and $d$ is the total number of features. Each element of $\boldsymbol{C}$ is defined as $C_{ij}=\frac{\sum_b z_{i,b}z'_{j,b}}{\sqrt{\sum_b(z_{i,b})^2}\sqrt{\sum_b(z'_{i,b})^2}}$, where $z_i$ and $z_i'$ are mean centred.
While in \cite{zbontar2021barlow} the cross-covariance matrix is computed considering the original images and their augmented versions, in this paper we propose to re-purpose it for domain alignment using corresponding samples across the two domains.
The loss in Eq.\ref{eq:IB} is a trade-off between two objectives, the first term that pushes the learned representation to be domain invariant and a second term that decorrelates the different components of the embedding. 
To build $\boldsymbol{C}$, we introduce a projection head $P$, similar to the one in \cite{zbontar2021barlow}, mapping $f^S$ and $f^T$ to $z^S$ and $z^T$.
Then, each source instance representation $z_i^S$ is paired with all target instance representations $z_j^T$ where the label of instance $i$ and the pseudo-label of instance $j$ are equal.
Note that the same instance $i$ or $j$ can appear in more than one pair.
We also introduced a queue $Q$ to keep recent $z^S$, effectively increasing the number of possible instances that are paired with $z^T$ in the minibatch.
After forming this list of pairs, the cross-correlation matrix can be computed between the source instances and the target instances of all pairs.
This process makes the model invariant to instances of different domains and enables tackling the domain adaptation setting. Our final loss, introducing a weighting factor $\alpha$, is then defined as follows:
\begin{equation}
\begin{aligned}
\mathcal{L} = L^{\mathcal{CE}} + \alpha \, \mathcal{L}^{\mathcal{IB}}.
\end{aligned}
\end{equation}

\section{Experiments}
\label{experiments}

\noindent \textbf{Datasets.} We conduct an extensive evaluation of UDAVT on two benchmarks for UDA in action recognition, namely \textit{HMDB$\leftrightarrow$UCF} \cite{chen2019temporal} and \textit{Kinetics$\rightarrow$NEC-Drone} \cite{choi2020unsupervised}. The former setting comprises videos from the \textit{HMDB51} \cite{kuehne2011hmdb} and \textit{UCF101} \cite{soomro2012ucf101} action recognition datasets, which both contain real videos downloaded from Youtube. In this case, the domain shift is therefore present, but limited. \textit{Kinetics$\rightarrow$NEC-Drone}, consists of videos from the large scale \textit{Kinetics} dataset \cite{kinetics-dataset}, that contains sequences from Youtube, and the \textit{NEC-Drone} \cite{choi2020unsupervised} dataset, which consists of video sequences taken from moving drones in an indoor environment. Furthermore, the video sequences of \textit{NEC-Drone} comprise high-resolution frames (1920x1080), and the action is often relegated to the corner of the frame and in many cases, the view is extremely slanted. Understandably, this setting is characterised by a significantly more challenging domain shift that consequently induces all the tested state-of-the-art methods to perform poorly on the original data. To alleviate this problem, we employed a pre-processing step exploiting a pretrained YOLO-based \cite{redmon2018yolov3} human detection model using AlphaPose \cite{fang2017rmpe} to identify and locate the human actor(s) and then crop around the humans with a minimal resolution of 224x224. Table \ref{tab:cropping_results} reports an overview of the boost observed when applying different models to the cropped version of the dataset. 
As it can be seen, our pre-processing enables a consistent and significant improvement in performance for all methods, with the gain ranging from 14\% to 50\%. The Table also includes the performance of SAVA \cite{Choi2020ShuffleAA}, which proposed by the researchers who originally released the dataset\footnote{The original code for SAVA was not available so the result for the cropped version of \textit{Kinetics$\rightarrow$NEC-Drone} was obtained with our implementation.}, and our best competitor CO$^2$A \cite{da_Costa_2022_WACV}. In the following experiments, we used the cropped version of the 
\textit{NEC-Drone} dataset.

\begin{table}[t]
    \caption{Effect of pre-processing on accuracy for \textit{Kinetics$\rightarrow$NEC-Drone}}
    \centering
    \resizebox{0.48\textwidth}{!}{
        \begin{tabular}{l|c|c}
            \toprule
            \textbf{Method} & \textbf{Original data} & \textbf{Cropped data} \\
            \midrule
            Source only & 15.0 & 29.4 (+14.9) \\
            SAVA \cite{Choi2020ShuffleAA} & 31.6 & 42.5  (+10.9) \\
            CO$^2$A \cite{da_Costa_2022_WACV} & 33.2 & 47.9 (+14.7) \\
            UDAVT & 16.0 & \textbf{65.3 (+49.3)} \\
            UDAVT - \textit{supervised}   & \textit{30.0} & \textit{78.9 (+48.9)} \\
            \bottomrule
        \end{tabular}
    }
    \vspace{-12pt}
    \label{tab:cropping_results}
\end{table}

\noindent \textbf{Implementation details.} Our method was implemented using Pytorch \cite{NEURIPS2019_9015}. The model was fine-tuned in phase 1 starting from a STAM model pretrained on Kinetics-400.
Subsequently, the partial fine-tune was carried out for 20 epochs with SGD, cosine learning rate decay, learning rate $0.001$, weight decay of $1e^{-9}$ and batch size 8.
In the second phase, the model was trained for an additional 20 epochs using the same optimiser, learning rate scheduler and weight decay, however, with a learning rate of $0.005$ and a batch size of $64$ instances per domain.
Additionally, the projection head $P$ was not trained, as we found that using a fixed random non-linear projection made the model more resilient to bad pseudo-labels.
Since this results in fewer parameters to optimise, we hypothesise that this makes training more consistent.
For HMDB$\leftrightarrow$UCF, we set $\alpha = 0.01$ and the queue size to $1024$.
For Kinetics$\rightarrow$NEC-Drone, we used $\alpha = 0.025$ and a queue of size $2048$. 
First, $\alpha$ needs to be set to a small value due to the sheer difference in magnitude between $\mathcal{L}^\mathcal{CE}$ and $\mathcal{L}^\mathcal{IB}$.
Second, $\alpha$ controls the strength of decorrelating feature representations of two different instances across domains that share the same label (pseudo-label for the target domain). Intuitively, it needs to be set higher the larger the domain gap is.

\noindent \textbf{Baselines.} We compare our results with those obtained by state-of-the-art methods for UDA in video action recognition, namely TA$^3$N \cite{chen2019temporal}, TCoN \cite{pan2020adversarial}, SAVA \cite{Choi2020ShuffleAA} and CO$^2$A \cite{da_Costa_2022_WACV}. For a fair comparison, the results of TA$^3$N are also reported after replacing their ResNet backbone with I3D. Also, as a transformer-based baseline, we report the results obtained by replacing our proposed UDAVT loss with three different domain alignment strategies, namely: a Maximum Mean Discrepancy (MMD) domain alignment component \cite{long2015learning}, an adversarial approach relying on a domain classifier as in \cite{ganin2015unsupervised} and a Maximum Classifier Discrepancy (MCD) based component \cite{saito2018maximum}. MCD aligns domains by employing task-specific decision boundaries that maximise the discrepancy between the output of two distinct classifiers to detect target samples lying far from source support and minimise the discrepancy of the transformer, so it learns how to produce target features closer to source support. The adversarial-based approach \cite{ganin2015unsupervised} consists of adding an MLP-based domain classifier that is responsible for predicting the domains of the instances given their video-level feature representations. We added a target cross entropy based on the pseudo-labels to all baselines was we found that this improved performance.

\subsection{Results}

Table \ref{tab:results_h&u} presents the results on \textit{HMDB$\leftrightarrow$UCF}. Along with the scores achieved with our proposed method, we report the ones obtained by previous approaches in the same settings. As it can be observed, all transformer-based models significantly outperform previous methods (except for CO$^2$A \cite{da_Costa_2022_WACV}) in both directions, suggesting that transformer-based methods are more robust to domain shift even without any domain adaptation strategy. In particular, we achieve an accuracy of 96.8\% and 92.3\% in the two directions, outperforming the current best competitor (CO$^2$A \cite{da_Costa_2022_WACV}) by 1\% and 4.5\%, respectively. Results also show that our method outperforms MMD with a transformer-based architecture. Also, the proposed MCD and adversarial-based baselines are outperformed (in just one of the two directions for the case of MCD). Finally, we report, as upper bounds, the scores obtained with the supervised version of UDAVT, i.e., the case where ground truth target labels are used instead of pseudo-labels to compute the cross-correlation matrix.
Table \ref{tab:results_k2n} reports the scores obtained on the \textit{Kinetics$\rightarrow$NEC-Drone} benchmark. This setting corresponds to a more significant domain shift since the target video sequences are shot by drones in a specific indoor environment. For this reason, it is easy to observe that the absolute value of all the reported scores is significantly lower when compared to the accuracy obtained in the previous benchmarks. However, the results clearly show how the transformer-based approaches strongly outperform the baselines achieving a score of 65.3\%, which is about 17 points more than the best competitor. In addition, the proposed loss achieves more than 10 points when compared to the MMD-based transformer. The gap is wider when it comes to the MCD and adversarial-based baselines, which are outperformed by 27 and 25 points. These experiments show that (i) the transformer-based backbone proves effective when applied to cases where a higher domain shift is present and (ii) the proposed UDAVT alignment method addresses the domain gap more efficiently leading to a significant increase in accuracy on the target domain.

\begin{table}[t]
    \centering
    \caption{Results on \textit{HMDB$\leftrightarrow$UCF}.}
    \resizebox{0.46\textwidth}{!}{
        \begin{tabular}{l|c|c|c}
        \toprule
        \textbf{Method} & \textbf{Encoder} & \textbf{H$\rightarrow$U} & \textbf{U$\rightarrow$H} \\
        \midrule
        \multicolumn{4}{c}{\textbf{Baselines}} \\
        \midrule
        Source only \cite{chen2019temporal} & \multirow{7}{*}{ResNet} & 71.7 & 73.9 \\
        DANN \cite{ganin2014unsupervised} & & 76.3 & 75.2 \\
        JAN \cite{long2017deep} & & 74.7 & 79.6 \\
        AdaBN \cite{li2016revisiting} & & 72.2 & 77.4 \\
        MCD \cite{saito2018maximum} & & 73.8 & 79.3 \\
        TA$^3$N \cite{chen2019temporal} & & 81.8 & 78.3 \\
       {Target only} \cite{chen2019temporal} & & \textit{82.8} & \textit{94.9} \\
        \midrule
        TCoN \cite{pan2020adversarial} & 2D/3D CNN & 89.1 & 87.2 \\ 
        \midrule
        Source only \cite{Choi2020ShuffleAA} & \multirow{4}{*}{I3D} & 88.8   & 80.3   \\
        TA$^3$N \cite{chen2019temporal} & & 90.5 & 81.4 \\
        SAVA \cite{Choi2020ShuffleAA} &  & 91.2 & 82.2 \\
        CO$^2$A \cite{da_Costa_2022_WACV} & & 95.8 & 87.8 \\
        {Target only} \cite{Choi2020ShuffleAA}&  & \textit{95.0}  & \textit{96.8}   \\
        \midrule
        \multicolumn{4}{c}{\textbf{Transformer-based}} \\
        \midrule
        Source only & \multirow{7}{*}{Transformer} & 93.7 & 86.9 \\
        MMD \cite{long2015learning} & & 96.5 & 87.9 \\
        MCD \cite{saito2018maximum} & & \textbf{97.2} & 87.9 \\
        Adversarial \cite{ganin2015unsupervised}  & & 96.6 & 87.6 \\
        UDAVT (ours) & & 96.8 & \textbf{92.3} \\
        UDAVT (ours) - \textit{supervised} & & \textit{97.2} & \textit{94.4 }\\
        {Target only} & & \textit{97.9} & \textit{95.8} \\
        \bottomrule
        \end{tabular}
    \label{tab:results_h&u}
}
\vspace{-12pt}
\end{table}

\begin{table}[t]
    \centering
    \caption{Results on \textit{Kinetics$\rightarrow$NEC-Drone}.}
    \resizebox{0.46\textwidth}{!}{
        \begin{tabular}{l|c|c}
            \toprule
            \textbf{Method} & \textbf{Encoder} & \textbf{Top-1 Acc} \\
            \midrule
            \multicolumn{3}{c}{\textbf{Baselines}} \\
            \midrule
            Source only & \multirow{2}{*}{Resnet} & 15.8 \\
            TA$^3$N \cite{chen2019temporal} & & 28.0 \\
            \midrule
            Source only & \multirow{3}{*}{I3D} & 32.0 \\
            TA$^3$N \cite{chen2019temporal} & & 44.7 \\
            SAVA \cite{Choi2020ShuffleAA} & & 42.5 \\
            CO$^2$A \cite{da_Costa_2022_WACV} & & 45.8 \\
            \midrule
            \multicolumn{3}{c}{\textbf{Transformer-based}} \\
            \midrule
            Source only & \multirow{7}{*}{Transformer} & 29.4 \\
            MMD \cite{long2015learning} & & 54.4 \\
            MCD \cite{saito2018maximum} & & 38.1 \\
            Adversarial \cite{ganin2015unsupervised} & & 40.8 \\
            UDAVT (ours) & & \textbf{65.3} \\
            UDAVT (ours) - \textit{supervised} & & \textit{78.1} \\
            \textit{Target only} & & \textit{82.9} \\
            \bottomrule
        \end{tabular}
    }
    \label{tab:results_k2n}
\vspace{-12pt}
\end{table}

\begin{table}[t]
    \centering
    \caption{Application of different self-supervised methods for domain alignment.}
    \resizebox{0.46\textwidth}{!}{
        \begin{tabular}{l|c|c|c|c}
        \toprule
        \textbf{Method} & \textbf{Encoder} & \textbf{H$\rightarrow$U} &
        \textbf{U$\rightarrow$H} & \textbf{K$\rightarrow$N-D} \\
        \midrule
        Source only & \multirow{4}{*}{Transformer} & 93.7 & 86.9 & 29.4 \\
        SimCLR \cite{chen2020simple} & & 95.4 & 85.8 & 38.6 \\
        VICReg \cite{bardes2021vicreg} & & 95.1 & 86.4 & 46.5 \\
        UDAVT (ours) & & \textbf{96.8} & \textbf{92.3} & \textbf{65.3} \\
        \bottomrule
        \end{tabular}
    }
    \label{tab:ss_results}
\vspace{-12pt}
\end{table}

One question that might arise is: what if the pseudo-labels have poor quality?
We want to show that the domains are aligned even if the pseudo-labels are initially causing clusters of different classes to form.
Since UDAVT is strongly inspired by the IB principle, commonly employed for self-supervised learning, we also perform an additional evaluation re-purposing other methods, i.e., SimCLR \cite{chen2020simple} and VICReg \cite{bardes2021vicreg}, for domain adaptation. 
Both methods have been proposed for unsupervised representation learning and, to adapt them for UDA, we used the same procedure employed to construct pairs in UDAVT, i.e., we replaced distorted versions of the same instances with pairs of source and target instances with corresponding label/pseudo-label.
VICReg is similar to Barlow Twins but explicitly avoids the collapsing problem. It employs different terms to maintain invariance to data augmentation, diversifies feature representations for different data, and enforces different features to encode different information. SimCLR relies on the concept of positive and negative pairs and, by using the InfoNCE loss \cite{oord2019representation}, it performs an optimisation procedure that consists of pulling the feature representations of positives closer together whereas those of the negatives are pushed farther away. We adapted this loss for DA using a similar formulation to \cite{khosla2021supervised}, in which multiple positives can be considered for the same instance. In the proposed model, positive instances are those from different domains that share the same label or pseudo-label and negatives are instances from a different domain, but without matching labels. To avoid specialising within a domain, we did not consider inter-domain pairs as neither negatives nor positives. Results for these losses, compared to our UDAVT, are reported in Table \ref{tab:ss_results}. These results show that although re-purposing the considered self-supervised methods to DA produces competitive performance in the target domain, our proposed UDAVT loss outperforms both strategies. In particular, in the challenging \textit{Kinetics$\rightarrow$NEC-Drone} setting the performance boost is more than 16\% higher than the second-best method.

\begin{figure}[t]
    \centering
    \includegraphics[width=0.4\textwidth]{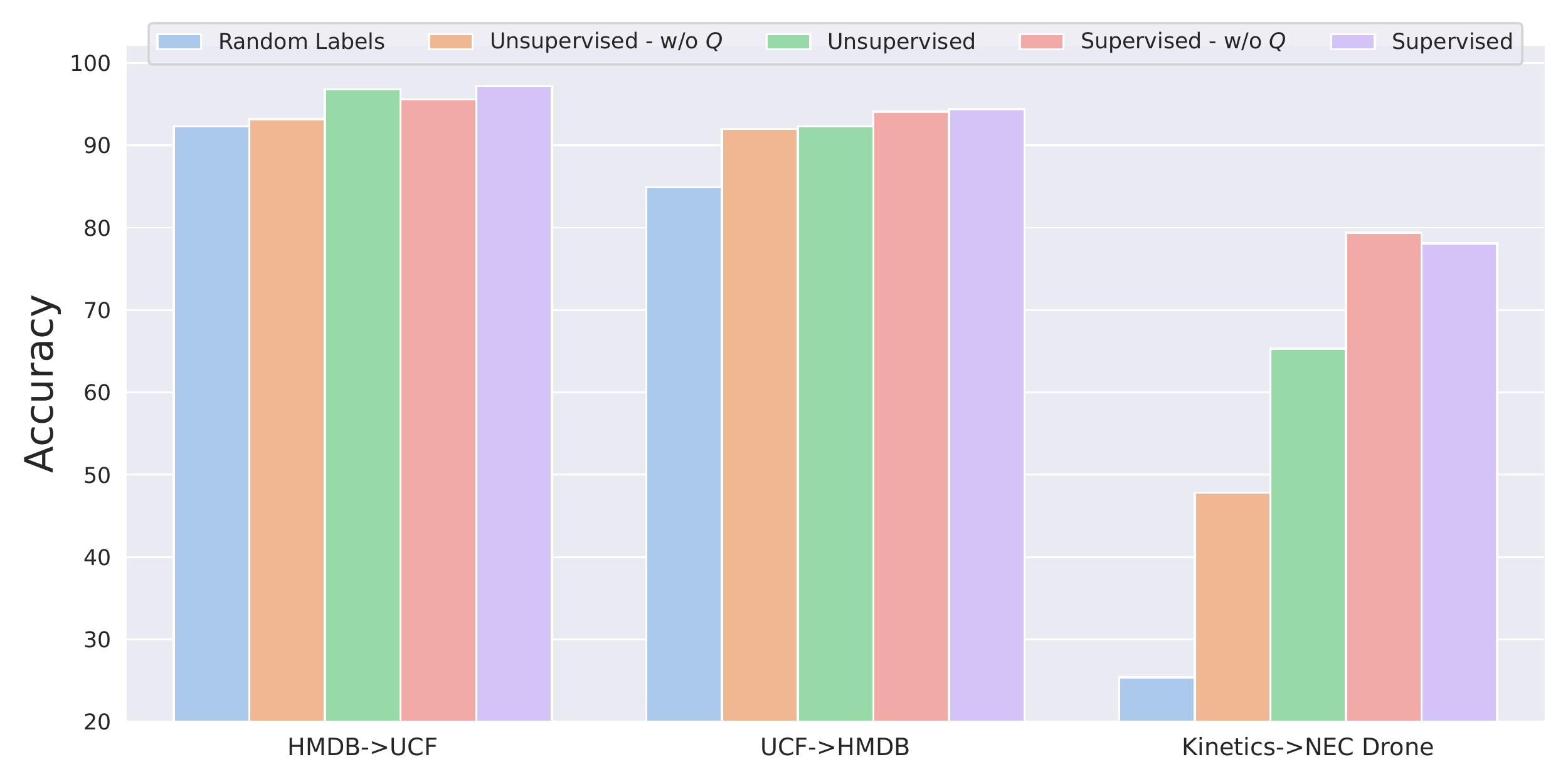}
    \vspace{-10pt}
    \caption{Ablation on the usage of the queue $Q$ and random labels for target data instead of pseudo-labels for domain alignment.}
    \label{fig:ablation}
    \vspace{-12pt}
\end{figure}

Lastly, we report in Fig. \ref{fig:ablation}, the ablation study that we carry out to study the performance of the model concerning (i) the usage of the queue $Q$ and (ii) the usage of target pseudo-labels for the domain alignment component. For the first, we removed $Q$, while for the latter we used random labels for target data. We can observe that the removal of $Q$ is detrimental to the overall performance of the model in the supervised and unsupervised cases, respectively, except for the supervised setting of \textit{Kinetics$\rightarrow$NEC-Drone}, where the model behaves very similarly. Regarding the labels, results show that the model always benefits from being provided with better target labels.

\section{Conclusions}
\label{conclusions}

We presented UDAVT, a simple and novel framework for unsupervised domain adaptation in video action recognition, which couples a powerful transformer-based feature extractor with a domain alignment component that exploits the Information Bottleneck principle to perform domain alignment. We reported results on two popular benchmarks for domain adaptation in action recognition, proving the effectiveness of UDAVT by outperforming previous state-of-the-art models in all settings, and further provided insight by exploring self-supervised contrastive variants for domain alignment, all of which proved effective, yet inferior to our proposed IB loss. In future work, we plan on improving our approach by further leveraging the capabilities of visual transformers for video action recognition while devising more sophisticated domain alignment strategies accordingly. Another research direction would be adapting the model to open set domain adaptation.


\textbf{Acknowledgment:} This work was supported by the Huawei Technologies project BONSAI, the EU H2020 AI4Media No. 951911 project, the EUREGIO project OLIVER and the National Council for Scientific and Technological Development (CNPq), Brazil and by the EU H2020 SPRING No. 871245 Projects.




\bibliographystyle{IEEEtran}
\bibliography{IEEEabrv,IEEEexample}

\end{document}